\documentclass[conference]{IEEEtran}
\usepackage{amsmath,amssymb,amsfonts}
\usepackage{acronym}
\usepackage{siunitx}
\usepackage{algpseudocode}
\usepackage{algorithm}
\usepackage{graphicx}
\usepackage{textcomp}
\usepackage{diffcoeff}
\usepackage{caption}
\usepackage{blindtext}
\usepackage{subcaption}
\usepackage{xcolor}

\usepackage{hyperref}
\hypersetup{
	colorlinks,
	linkcolor={red!50!black},
	citecolor={blue!50!black},
	urlcolor={blue!80!black}
}

\usepackage{orcidlink}                                  

\definecolor{orcidlogocol}{HTML}{A6CE39}

\acrodef{PDE}{Partial Differential Equation}
\acrodef{FDM}{Finite Difference Method}
\acrodef{pdf}{probability density function}
\acrodef{SBL}{sparse Bayesian learning}
\acrodef{ARD}{Automatic Relevance determination}
\acrodef{R-ARD}{reformulated Automatic Relevance determination}
\acrodef{IR-ARD}{incremental reformulated Automatic Relevance determination}
\acrodef{GSL}{Gas Source Localization}
\acrodef{RHS}{right-hand side}
\acrodef{FMLM}{Fast Marginal Likelihood Maximization}
\acrodef{CTA}{combine-then-adapt}
\acrodef{PINN}{Physics-Inspired Neural Network}
\acrodef{MLP}{Multilayer Perceptron}
\acrodef{NN}{Neural Network}
\acrodef{FEM}{Finite Element Method}
\acrodef{BC}{boundary condition}
\acrodef{PGNN}{Physics-Guided Neural Network}
\acrodef{MSE}{Mean Square Error}
\newcommand{\orcid}[1]{\href{https://orcid.org/#1}{\textcolor[HTML]{A6CE39}{\aiOrcid}}}

\newcommand*{\vect}[1]{\boldsymbol{#1}}

\newcounter{remarknumber}
\usepackage[url=false,style=ieee, isbn=false, doi=false,eprint=false,maxnames=3,minnames=1]{biblatex}

\IEEEoverridecommandlockouts  
\begin{document}

\title{Gas Source Localization Using Physics-Guided Neural Networks}



\author{
\IEEEauthorblockN{
Victor 
Prieto Ruiz, 
Patrick Hinsen,\\ 
Thomas Wiedemann, 
and
Dmitriy Shutin 
}
\IEEEauthorblockA{
\emph{Institute of Communications and
Navigation,}\\ 
\emph{German Aerospace Center (DLR),}
Oberpfaffenhofen, Germany 
}\\[-0.8cm]
\and
\IEEEauthorblockN{Constantin Christof}
\IEEEauthorblockA{
\emph{School of Computation,
Information and Technology},\\
\emph{Technical University of Munich,}
\\
Garching, Germany
}\\[-0.8cm]
\thanks{ 
This work was supported by the EU Project TEMA. 
The TEMA  project has received funding from the European Commission under HORIZON EUROPE (HORIZON Research and Innovation Actions) under grant agreement 101093003 (HORIZON-CL4-2022-DATA-01-01). Views and opinions expressed are however those of the author(s) only and do not necessarily reflect those of the European Union or Commission. Neither the European Commission nor the European Union can be held responsible for them.}
}
\maketitle

\begin{abstract}

This work discusses a novel method for estimating the location of a gas source based on spatially distributed concentration measurements taken, e.g., by a mobile robot or flying platform that follows a predefined trajectory to collect samples.
The proposed approach uses a Physics-Guided Neural Network to approximate the gas dispersion with the source location as an additional network input.
After an initial offline training phase, the neural network can be used to efficiently solve the inverse problem of localizing the gas source based on measurements. 
The proposed approach allows avoiding rather costly numerical simulations of gas physics needed for solving inverse problems.
Our experiments  show that the method localizes the source well, even when dealing with measurements affected by noise.  
\end{abstract}


\section{Introduction}

\label{sec:introduction}

The problem of \ac{GSL}
(e.g., of a gas leak) by means of measurements taken by mobile agents has received increasing attention in the last decades due to rapid improvements in robotics and gas sensor technology. 
While biologically inspired localization approaches, such as chemotaxis or anemotaxis, have initially served as simple guiding principles, these strategies have been found to be not robust enough for efficient source localization \cite[Sec.~4.1]{Francis2022}.
More recent probabilistic model-based approaches, such as infotaxis, have proved more reliable in patchy and turbulent regimes.
However, they rely on assumptions on the source distribution, making them highly dependent on prior knowledge \cite[Sec.~6]{Jing2021}. 
Further, the incorporated models often come with high computational costs. 
For a detailed review of different \ac{GSL}-approaches, 
we refer to \cite{Francis2022} and the references therein. 

In the present work, we propose a strategy which approximates the complex gas dispersion physics through the use of an easy-to-evaluate neural network. 
More precisely, a \ac{PGNN} \cite[Secs.~1,2]{Faroughi2022} is used as a surrogate model for the advection-diffusion \ac{PDE}.
This network is trained to learn and efficiently emulate the complicated functional dependency between the spatial gas concentration and the source location. 

Our surrogate model approach offers several advantages:
First, it is flexible and can be applied to various types of \acp{PDE} with different \acp{BC} and parameters.
Second, our surrogate model 
can be differentiated easily with respect to 
the source location, thanks to automatic differentiation libraries.
This makes the model extremely useful 
in the inverse problem setting that requires a gradient-based numerical optimization method.
Third, the computationally expensive
network training can be done once offline.
Afterwards, the network can be used for solving \ac{GSL}-problems with different observations, without the need for retraining or a costly \ac{PDE}-solver.  
This opens up the possibility for real-time applications, even those using hardware-constrained robots.
Lastly, our method 
is also robust to noisy measurements and, thus, 
able to handle inaccuracies caused by sensor limitations.

\section{Methodology}
\label{sec:GasModel} 
\subsection{Gas Model}
We assume that the gas occupies 
the spatial domain
$\Omega = (\SI{0}{\kilo\meter},\SI{1}{\kilo\meter})\times (\SI{0}{\kilo\meter},\SI{1}{\kilo\meter})$. 
On the set $\Omega$, we model the gas spread by the linear advection-diffusion \ac{PDE}. 
For the sake of simplicity,
we consider a time-invariant (equilibrium) problem with homogeneous Dirichlet \acp{BC}. This is given by the \ac{PDE} 
\begin{equation}
\label{eq:AdvecDiffuseEquil}
\begin{aligned}
\partial_t u(\vect x) = \kappa\Delta u(\vect x) - \vect v\cdot \nabla u(\vect x) + s(\vect x) &=0   &&\text{in } \Omega,\\
u(\vect x)&=0 &&\text{on } \partial\Omega.
\end{aligned}
\end{equation}
Here, $\partial\Omega$ is the boundary of $\Omega$, $u\colon \Omega \to \mathbb R$ is the gas concentration,
$s$ is a source term (either a
real-valued function or a measure on $\Omega$),
$\kappa= \SI{1}{\square\kilo\meter\per\hour}$ is the diffusion coefficient, and  $\vect v = [\SI{3}{\kilo\meter\per\hour},\SI{3}{\kilo\meter\per\hour}]^\top$ is the advection velocity.
We assume a uniform (known) wind flow.

\begin{figure}
    \centering
    \includegraphics[width=.7\linewidth]{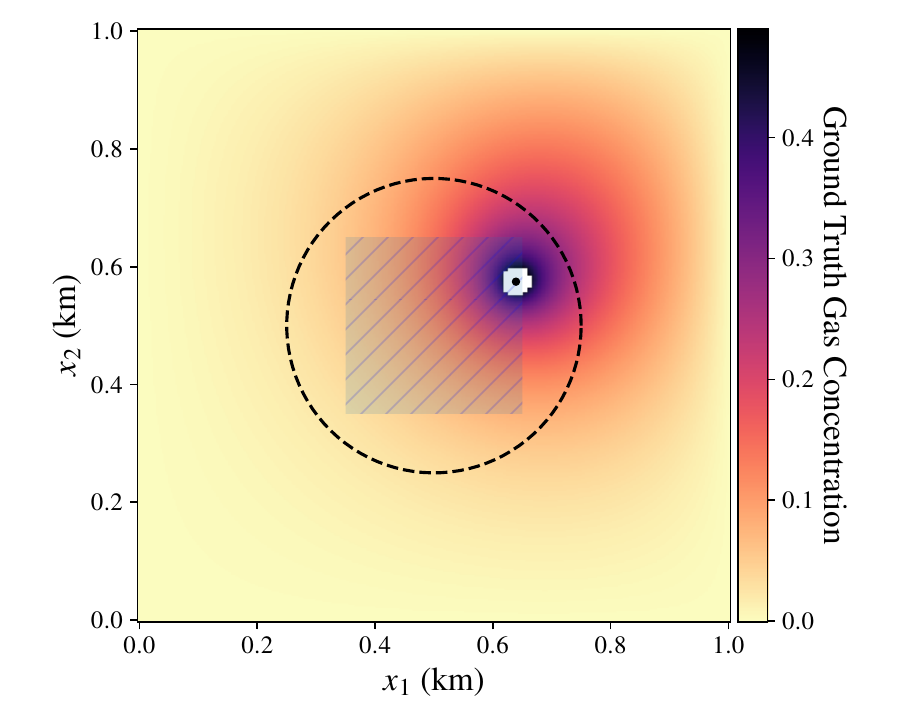}
    \vspace{-2mm}
    \caption{
    Problem setup. The gas concentration field is shown as an example. (A ball of radius 
    \SI{20}{\meter} around $\vect p$ is excluded to avoid the singularity.) The dotted black circle represents $D_\mathrm{O}$, the trajectory of the observations. The source location $\vect p$ is marked by a black dot and the set
    $\mathcal{P}$ of possible source positions by the blue hatched region.}
    \label{fig:ProbSetup}
    \vspace{-1.5mm}
\end{figure}

Since we are interested in \ac{GSL},  we assume that there is a location parameter set $\mathcal{P} = [\SI{.35}{\kilo\meter},\SI{.65}{\kilo\meter}]\times [\SI{.35}{\kilo\meter},\SI{.65}{\kilo\meter}] \subset \Omega$ which fully represents all possible points $\vect p \in \mathbb R^2$ where the source could be located.
We shall use $s_{\vect p}$ and 
$u_{\vect p}$ to denote the source term 
and the gas concentration field 
associated with the source position 
$\vect p$, respectively. That is, $u_{\vect p}$ is the solution of \eqref{eq:AdvecDiffuseEquil} with source term $s_{\vect p}$.

For the purposes of this paper, we restrict our attention to ideal point sources, i.e., Dirac delta distributions.
Keeping with previous notation, this means that we consider source terms of the form 
$s_{\vect p} := \delta_{\vect p}$, $\vect p \in \mathcal{P}$. Note that, for such 
$s_{\vect p}$, the solutions $u_{\vect p}$ of 
\eqref{eq:AdvecDiffuseEquil} possess a singularity at $\vect p$. 
This means in particular that 
$u_{\vect p}$ cannot be evaluated at 
$\vect p$.


\subsection{Source Localization Problem}
In a source localization problem governed by \eqref{eq:AdvecDiffuseEquil}, the goal is to identify the location $\vect p \in \mathcal{P}$ associated with the source term $s_{\vect p}$ from measurements of $u_{\vect p}$ taken by a mobile robot on an observation domain $D_\mathrm{O}\subset \Omega$. 
The set $D_\mathrm{O}$ can be considered as the 
robot trajectory
in the  region of interest;
cf.\ Fig.~\ref{fig:ProbSetup}.

We suppose that the gas concentration measurements are taken
at $N_\mathrm{O}$ observation points $\vect x_1,..., \vect x_{N_\mathrm{O}} \in D_\mathrm{O}$.
A measurement $y_j$ at
$\vect x_j$ is the gas concentration $u_{\vect p}(\vect x_j)$
perturbed by zero-mean additive Gaussian noise with standard deviation $\sigma$.
To avoid evaluating the solution 
of \eqref{eq:AdvecDiffuseEquil} at its singularity, 
we henceforth assume that $\mathcal{P}\cap D_\mathrm{O}=\emptyset$. We remark that this restriction can be avoided by replacing the point sources $\delta_{\vect p}$
with a suitable mollification.

To determine the source location based 
on measurements $y_1,...,y_{N_\mathrm{O}}$, we consider the minimization problem
\begin{equation}
    \label{eq:InvProb}
 \min_{\vect p'\in \mathcal{P}}~
 \mathcal{J}(\vect p') := \sum_{j=1}^{N_\mathrm{O}} (y_j-u_{\vect p'}(\vect x_j))^2.
\end{equation}
Note that standard optimization algorithms for  
\eqref{eq:InvProb} require evaluations of the 
objective $\mathcal{J}$ and its derivatives at numerous points $\vect p'$. 
Due to the definition of $u_{\vect p'}$, 
for each of these evaluations, 
one has to
solve 
the PDE \eqref{eq:AdvecDiffuseEquil} 
with parameter $\vect p'$
and, in the case that derivatives are required, an additional adjoint PDE. 
This is often computationally prohibitive,
in particular if the aim is to solve 
\eqref{eq:InvProb} with limited hardware, e.g., directly on-board a robot.
The main idea of 
our approach is to resolve this problem
by setting up a reusable 
neural network model for 
the functional dependency 
$(\vect x, \vect p) \mapsto u_{\vect p}(\vect x)$ which is trained once offline, i.e., 
we use a surrogate for $u_{\vect p'}$ in \eqref{eq:InvProb}.  


\subsection{\texorpdfstring{%
\ac{NN} Surrogate}{}}
\label{sec:NN}


\subsubsection{Network Architecture}

As our surrogate model for the functional dependency $(\vect x, \vect p) \mapsto u_{\vect p}(\vect x)$, 
we consider a \ac{MLP}, a fully-connected feed-forward neural network.
A popular architecture for an $L$-layer \ac{MLP} 
$\mathcal{N}\colon \mathbb{R}^{n_\text{in}} \to \mathbb{R}^{n_\text{out}}$
with input variable $\vect{z}$
is the following: 
\begin{equation}
\label{eq:network_def}
    \begin{aligned}
    \vect{l}_0 :=\vect{z},\quad
    \vect{l}_k &:=\rho(\vect{W}_k \vect{l}_{k-1} + \vect{b}_k),\,\, k = 1,...,L-1,\\
    \mathcal{N}(\vect{z})&:= \vect{W}_L\vect{l}_{L-1} + \vect{b}_L.
    \end{aligned}
\end{equation}
Here, $\vect{W}_k \in \mathbb R^{n_k \times n_{k-1}}, \vect{b}_k \in \mathbb R^{n_k}$ are the $k$-th layer's weight matrix and bias vector, respectively,  with $n_0,...,n_L \in \mathbb{N}$ the number of neurons at each layer ($n_0 = n_\text{in}$, $n_L = n_\text{out}$), and $\rho\colon \mathbb R\to \mathbb R$ is the nonlinear activation function, evaluated componentwise when applied to a vector. We note that the definition \eqref{eq:network_def} follows common practice and has the last network layer as a scaling layer without an activation function.

\begin{figure}
    \centering
    \includegraphics[width=.9\linewidth]{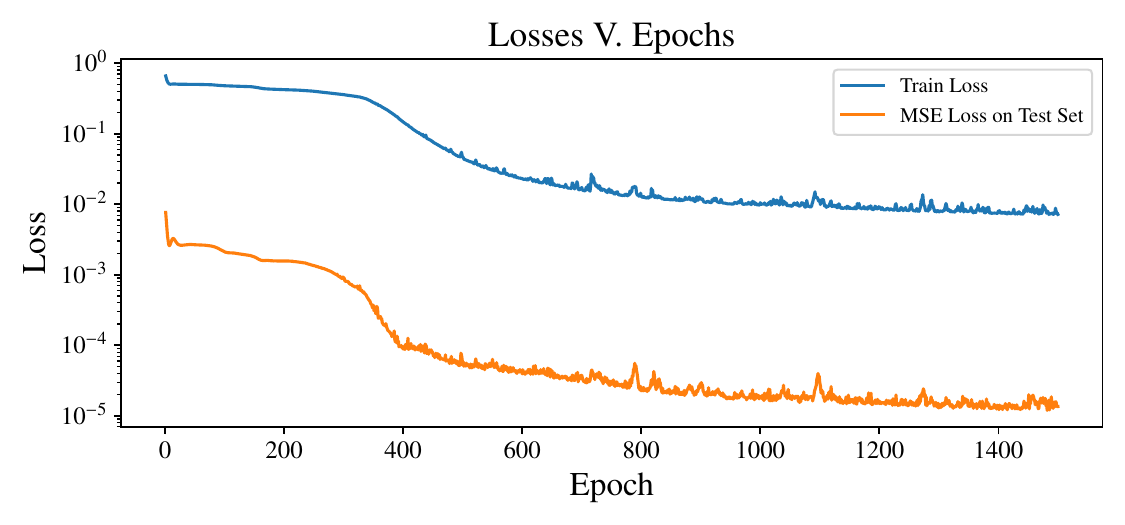}\vspace{-2mm}
    \caption{
    The surrogate model's performance over its training period. Blue: $H_1$-loss on training set. Orange: Mean Square Error (MSE) of function values on test set (withheld from NN-training).
    }
    \label{fig:NNTraining}
    \vspace{-1.5mm}
\end{figure}

In our application, 
the network concatenates both the spatial coordinate~$\vect x \in \Omega$ and the source location~$\vect p\in \mathcal{P}$ as a single input $\vect z := [\vect x; \vect p]$, and it outputs an approximation of  $u_{\vect p}(\vect x)$. This means $n_\text{in} = n_0 = 2 + 2 = 4, n_\text{out} = n_L = 1$. 


For the numerical experiments in this paper, we used a
network with 5 layers, 30 neurons in each hidden layer ($n_1,...,n_4 = 30$), 
and the {\tt{softplus}} activation function.

\subsubsection{Enforcing Boundary Conditions}
We use the hard-enforcement approach suggested in \cite{Sukumar2022} to realize the \acp{BC}. 
Our approximator multiplies the \ac{MLP} output $\mathcal{N}$ from \eqref{eq:network_def} by a cutoff function $\psi$ which vanishes on the spatial boundary $\partial \Omega$. This ensures homogeneous Dirichlet \acp{BC}. Thus, we define
\begin{equation}
\label{eq:tilde_u_def}
    \tilde{u}_{\vect p}(\vect x):= \psi(\vect x)\mathcal{N}([\vect x;\vect p]).
\end{equation}
As the function $\psi$, we use the so-called ``$R$-$m$ approximate distance function'' to the boundary $\partial\Omega$  \cite{Sukumar2022}. 

\begin{figure*}[!t]
\vspace{-3mm}
\centering
\begin{subfigure}[]{.42\linewidth}
    \centering
    \includegraphics[width=.8\linewidth]{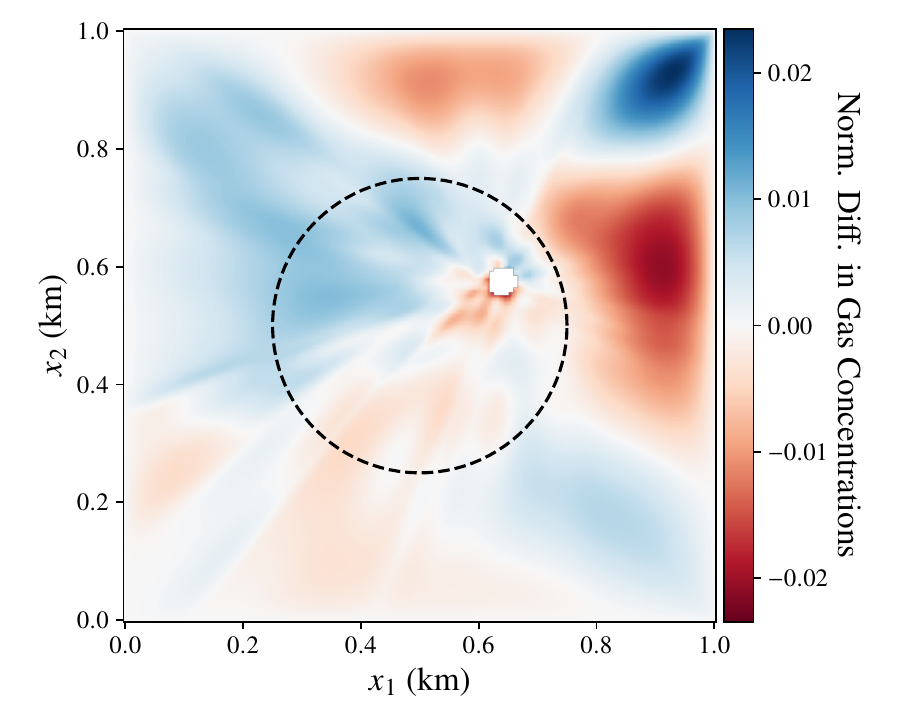}
    \vspace{-2.5mm}
    \caption{}
    \vspace{1.5mm}
    \label{fig:NNDiffPlot}
\end{subfigure}
\hspace{.05\linewidth}
\begin{subfigure}[]{.42\linewidth}
    \centering
    \includegraphics[width=.78\linewidth]{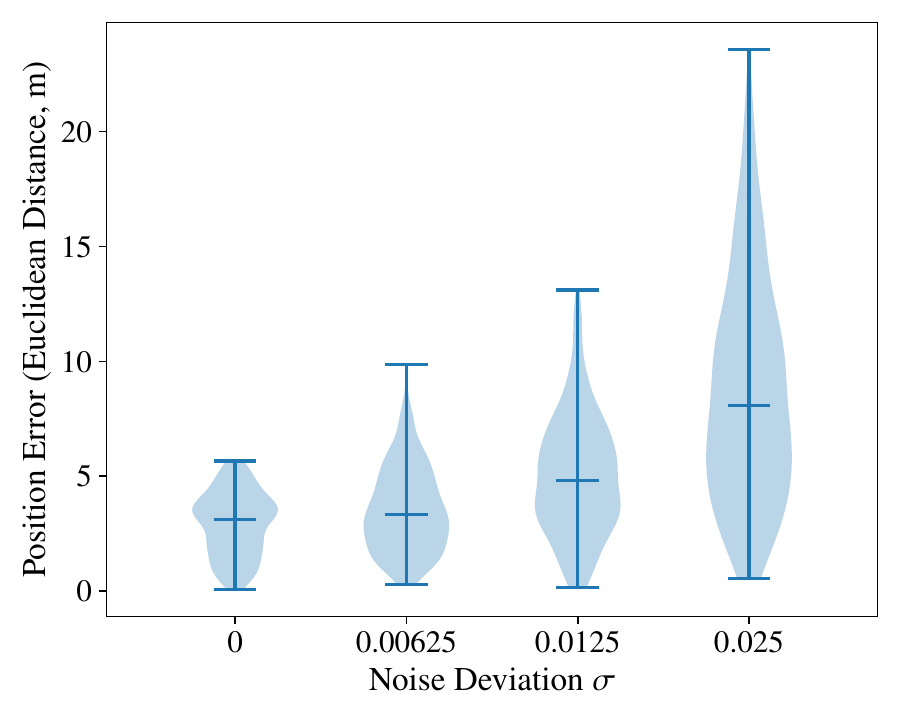}
    \vspace{-2.5mm}
    \caption{}
    \vspace{1.5mm}
    \label{fig:ViolinPlot}
    \end{subfigure}
    
\caption{\ac{NN} training results. Fig.\ (a) shows the difference between the 
NN-surrogate $\tilde u_{\vect p}$ and the reference solution $ u_{\vect p}$ in $\Omega$, normalized by the maximum value of $u_{\vect p}$ excluding a ball of radius \SI{20}{\meter} around the source, in the situation of Fig.~\ref{fig:ProbSetup}.
Fig.\ (b) shows the (Euclidean) position error of the 
location $\vect p$
calculated by solving 
\eqref{eq:InvProb} 
with our NN-surrogate approach 
with increasing amounts of White Gaussian Noise.
\vspace{-3mm}}
\label{fig:ResFigs}

\end{figure*}

\subsubsection{Network Training}
To train the NN-surrogate $\tilde u_{\vect p}$, we use a physics-guided approach. 
This supervised learning method compares the network output and its spatial gradients with a reference solution 
and attempts to make them as similar as possible.
We thus define the 
\emph{Physics-Guided 
$H_1$-Loss} 
by
\vspace{-0.1cm}
\begin{equation*}
    L_{\textbf{PG}} :=  \frac 1{N_t}
    \sum_{i=1}^{N_t}\left[(\tilde{u}_{\vect p_i} - u_{\vect p_i})(\vect{q_i})\right]^2 + \lVert (\nabla\tilde{u}_{\vect p_i} - \nabla u_{\vect p_i})(\vect{q_i}) \rVert_2^2.\vspace{-0.1cm}
\end{equation*}
Here, $\tilde u_{\vect p} $ 
is the model in \eqref{eq:tilde_u_def}, $u_{\vect p}$ denotes a reference solution computed, e.g., by the \ac{FEM}, $\nabla$ is the spatial gradient, 
$\{(\vect p_i, \vect q_i)\} \subset \mathcal{P} \times \Omega$ is a set of $N_t$ evaluation points, and $ L_{\textbf{PG}}$ is understood as a function of the weights and biases in $\mathcal{N}$.
Note that
the $H_1$-loss, as opposed to the naive \ac{MSE}, includes information about 
the gradient of the reference solution. 

In our numerical experiments, 
we used $N_t$ = 80000 points $(\vect p_i,\vect q_i)$ in  $L_{\textbf{PG}}$, sampled randomly from 
$\mathcal{P}\times \Omega$. Since $u_{\vect p}$ 
has a singularity at $\vect p$, we excluded points $\vect q_i$ with a distance smaller than \SI{0.02}{\kilo\meter} from the source location $\vect p_i$. 
For the generation of the reference data, we used a 
 discretization
based on the FEniCS library \cite{Alnaes2015}
on a Friedrichs--Keller triangulation on a $ 241\times 241$ grid and linear Lagrangian elements.
The NN-surrogate was trained and implemented in Python using the Pytorch library \cite{Paszke2019}. We used the ADAM Optimizer 
with learning rate $l=10^{-3}$ to train $\tilde u_{\vect p}$ over 1500 epochs of the dataset. We additionally computed the \ac{MSE} on another set of 20000 test points randomly sampled from $\mathcal{P} \times \Omega$, excluding singularities, to check the surrogate's generalization performance. 
The  training results can be seen in Fig.~\ref{fig:NNTraining}.

\subsection{Optimization Algorithm}

Once the NN-surrogate $\tilde u_{\vect p}$ has been trained, the inverse problem 
\eqref{eq:InvProb}, with 
$u_{\vect p}$ replaced by $\tilde u_{\vect p}$, can be solved with standard optimization
algorithms. In our experiments,
we used the LBFGS-B method \cite{Byrd1995} for this purpose. The method had information about the  gradients $\nabla_{\vect p} \tilde u_{\vect p}$ via the automatic differentiation functionalities of the 
Pytorch library \cite{Paszke2019}. 
Note that the optimization step
only requires gradients of the surrogate 
model $\tilde u_{\vect p}$. Measurements 
of the 
concentration gradients do not 
have to be taken in our \ac{GSL}-approach and we do not require that the actual concentration field is smooth. (It should, of course, be
such that it is sufficiently well described by \eqref{eq:AdvecDiffuseEquil}.)

\section{Results}\label{sec:Simulations}
\label{sec:Results}


To test our method's performance, we first checked that the surrogate model provides a sensible approximation. In addition to 
the loss metric in the training curve, we used a
comparison with a reference solution for this purpose. In Fig.~\ref{fig:NNDiffPlot}, we see the difference between modeled values and ground truth for a fixed source position, normalized by the maximum concentration value  of the reference function outside of its singular region. We can see that the difference is within 2\% of the maximum source value in all of $\Omega$.

Next, we tested the performance of
our \ac{PGNN} in the \ac{GSL}-problem \eqref{eq:InvProb}. To do this, we chose a set $\{\vect x_j\}$ of $N_\mathrm{O}=100$ evenly spaced points on the circle of radius \SI{0.25}{\kilo\meter} around
$(\SI{.5}{\kilo\meter},\SI{.5}{\kilo\meter})$ as the observation domain $D_\mathrm{O}$. 
Note that this choice of $D_\mathrm{O}$ is purely for demonstration purposes. More complicated $D_\mathrm{O}$ could be considered here 
without problems. 
We then sampled 324 source positions $\vect p$ inside the set
$\mathcal{P}$ on a grid. 
For each of these source positions, first, we used the \ac{FEM} to compute the ground truth values $y_j$ at the observation points $\vect x_j$ without noise, and then solved the inverse problem \eqref{eq:InvProb}, localizing the gas source to get the estimate $\hat {\vect p}$.
The source positioning error is measured as $\lVert \vect p-\hat{ \vect p}\rVert_2$.
We additionally repeated these tests, adding white Gaussian noise of standard deviation $\sigma\in\{0.00625, 0.0125, 0.025\}$ to the observations. Each inverse problem was solved in the span of 1 second.

The results of this survey 
are depicted in Fig.~\ref{fig:ViolinPlot} in a violin plot of the error. 
We can see that, without noise, the network can identify the correct source location up to an error of \SI{6}{\meter} in the worst case and up to an error of \SI{4}{\meter} on average. Given that the resolution of the \ac{FEM}-approximation is \SI{4.14}{\meter}, this performance is close to optimal. Even with additive noise corresponding to $\sim 6\%$ of the highest values encountered on the circle, the solver is still able to give source locations within \SI{28}{\meter} of the real value. We also notice that the median error (middle line) increases much slower than the maximum error.

\section{Conclusion}
In this paper, we have demonstrated how  \ac{PGNN}s can be used to solve
\ac{GSL}-problems in an accurate, computationally cheap, and robust manner.
Our approach is flexible and offers various  possibilities for extensions, e.g., to three dimensions and unknown 
wind/diffusion parameters. 
%
%
In future work, we plan 
to include dynamic path optimization techniques and realistic sensor models
into our framework. Further 
open points are the validation of our method
in real world experiments and the
detailed comparison with existing
\ac{GSL}-techniques---two issues that are beyond 
the scope of the present paper.

\printbibliography

\end{document}